\documentclass[11pt,a4paper]{article}
\usepackage[nohyperref]{latex/emnlp2018}
\usepackage{times}
\usepackage{latexsym}
\usepackage{natbib}
\usepackage{url}
\usepackage{graphicx}
\usepackage{xspace}

\newcommand{\lc}{\ensuremath{_\mathit{lc}}\xspace}
\newcommand{\moses}{{\sc{Moses}}\xspace}
\newcommand{\sacrebleu}{{\sc{SacreBLEU}}\xspace}
\newcommand{\sacrebleuversion}{1.2.10}
\newcommand{\unclear}{\emph{(unclear)}}

\title{A Call for Clarity in Reporting BLEU Scores}

\aclfinalcopy
\author{Matt Post \\
  Amazon Research \\
  Berlin, Germany}

\date{}

\begin{document}
\maketitle

\begin{abstract}
  The field of machine translation faces an under-recognized problem because of inconsistency in the reporting of scores from its dominant metric.
  Although people refer to ``the'' BLEU score, BLEU is in fact a parameterized metric whose values can vary wildly with changes to these parameters.
  These parameters are often not reported or are hard to find, and consequently, BLEU scores between papers cannot be directly compared.
  I quantify this variation, finding differences as high as 1.8 between commonly used configurations.
  The main culprit is different tokenization and normalization schemes applied to the reference.
  Pointing to the success of the parsing community, I suggest machine translation researchers settle upon the BLEU scheme used by the annual Conference on Machine Translation (WMT), which does not allow for user-supplied reference processing, and provide a new tool, \sacrebleu,\footnote{\url{https://github.com/awslabs/sockeye/tree/master/contrib/sacrebleu}} to facilitate this.
\end{abstract}

\section{Introduction}

Science is the process of formulating hypotheses, making predictions, and measuring their outcomes.
In machine translation research, the predictions are made by models whose development is the focus of the research, and the measurement, more often than not, is done via BLEU \citep{papineni2002:bleu}.
BLEU's relative language independence, its ease of computation, and its reasonable correlation with human judgments have led to its adoption as the dominant metric for machine translation research.
On the whole, it has been a boon to the community, providing a fast and cheap way for researchers to gauge the performance of their models.
Together with larger-scale controlled manual evaluations, BLEU has shepherded the field through a decade and a half of quality improvements \cite{graham2014:machine}.

This is of course not to claim there are no problems with BLEU.
Its weaknesses abound, and much has been written about them (cf.\ \citet{callison-burch2006:re-evaluation,reiter2018:structured}).
This paper is not, however, concerned with the shortcomings of BLEU as a proxy for human evaluation of quality; instead, our goal is to bring attention to the relatively narrower problem of the \emph{reporting} of BLEU scores.
This problem can be summarized as follows:
\begin{itemize}
\item BLEU is not a single metric, but requires a number of parameters (\S\ref{section:constellation}).
\item Preprocessing schemes 
  have a large effect on scores (\S\ref{section:processing}).
  Importantly, BLEU scores computed against differently-processed references are not comparable.
\item Papers vary in the hidden parameters and schemes they use, yet often do not report them (\S\ref{section:details}).
  Even when they do, it can be hard to discover the details.
\end{itemize}
Together, these issues make it difficult to evaluate and compare BLEU scores across papers, which impedes comparison and replication.
I quantify these issues and show that they are serious, with variances bigger than many reported gains.
After introducing the notion of \emph{user-} versus \emph{metric-supplied} tokenization, I identify user-supplied reference tokenization as the main cause of this incompatibility.
In response, I suggest the community use only \emph{metric-supplied} reference tokenization when sharing scores,\footnote{Sometimes referred to as \emph{detokenized BLEU}, since it requires that system output be detokenized prior to scoring.} following the annual Conference on Machine Translation \citep[WMT]{bojar2017:findings}.
In support of this, I release a Python package, \sacrebleu,\footnote{\texttt{pip3 install sacrebleu}} which automatically downloads and stores references for common test sets, thus introducing a ``protective layer'' between them and the user.
It also provides a number of other features, such as reporting a version string which records the parameters used and which can be included in published papers.

\begin{table*}[t]
  \centering\resizebox{\textwidth}{!}{
    \begin{tabular}{l|rrrrrr|rrrrrr}
      & \multicolumn{6}{c|}{English$\rightarrow\star$}       & \multicolumn{6}{|c}{$\star\rightarrow$English} \\
      config & en-cs & en-de & en-fi & en-lv & en-ru & en-tr & cs-en & de-en & fi-en & lv-en & ru-en & tr-en \\
      \hline\hline
      basic     & 20.7 & 25.8 & 22.2 & 16.9 & 33.3 & 18.5 & 26.8 & 31.2 & 26.6 & 21.1 & 36.4 & 24.4 \\
      split     & 20.7 & 26.1 & 22.6 & 17.0 & 33.3 & 18.7 & 26.9 & 31.7 & 26.9 & 21.3 & 36.7 & 24.7 \\
      unk       & 20.9 & 26.5 & 25.4 & 18.7 & 33.8 & 20.6 & 26.9 & 31.4 & 27.6 & 22.7 & 37.5 & 25.2 \\
      metric    & 20.1 & 26.6 & 22.0 & 17.9 & 32.0 & 19.9 & 27.4 & 33.0 & 27.6 & 22.0 & 36.9 & 25.6 \\
      \hline
      \emph{range}     &  0.6 &  0.8 &  0.6 &  1.0 &  1.3 &  1.4 &  0.6 &  1.8 &  1.0 &  0.9 &  0.5 &  1.2 \\
      \hline\hline
      basic\lc  & 21.2 & 26.3 & 22.5 & 17.4 & 33.3 & 18.9 & 27.7 & 32.5 & 27.5 & 22.0 & 37.3 & 25.2 \\
      split\lc  & 21.3 & 26.6 & 22.9 & 17.5 & 33.4 & 19.1 & 27.8 & 32.9 & 27.8 & 22.2 & 37.5 & 25.4 \\
      unk\lc    & 21.4 & 27.0 & 25.6 & 19.1 & 33.8 & 21.0 & 27.8 & 32.6 & 28.3 & 23.6 & 38.3 & 25.9 \\
      metric\lc & 20.6 & 27.2 & 22.4 & 18.5 & 32.8 & 20.4 & 28.4 & 34.2 & 28.5 & 23.0 & 37.8 & 26.4 \\
      \hline
      \emph{range}\lc  &  0.6 &  0.9 &  0.5 &  1.1 &  0.6 &  1.5 &  0.7 &  1.7 &  1.0 &  1.0 &  0.5 &  1.2 \\
    \end{tabular}
  }
  \caption{BLEU score variation across WMT'17 language arcs for cased (top) and uncased (bottom) BLEU.
    Each column varies the processing of the ``online-B'' system output and its references.
    \emph{basic} denotes basic user-supplied tokenization, \emph{split} adds compound splitting, \emph{unk} replaces words not appearing at least twice in the training data with UNK, and \emph{metric} denotes the metric-supplied tokenization used by WMT.
  The \emph{range} row lists the difference between the smallest and largest scores, excluding \emph{unk}.}
  \label{table:scores}
\end{table*}

\section{Problem Description}

\subsection{Problem: BLEU is underspecified}
\label{section:constellation}

``BLEU'' does not signify a single concrete method, but a constellation of parameterized methods.
Among these parameters are:
\begin{itemize}
\item The number of references used;
\item for multi-reference settings, the computation of the length penalty;
\item the maximum n-gram length; and
\item smoothing applied to 0-count n-grams.
\end{itemize}
Many of these are not common problems in practice.
Most often, there is only one reference, and the length penalty calculation is therefore moot.
The maximum n-gram length is virtually always set to four, and since BLEU is corpus level, it is rare that there are any zero counts.

But it is also true that people use BLEU scores as very rough guides to MT performance across test sets and languages (comparing, for example, translation performance into English from German and Chinese).
Apart from the wide intra-language scores between test sets, the number of references included with a test set has a large effect that is often not given enough attention.
For example, WMT 2017 includes two references for English--Finnish.
Scoring the online-B system with one reference produces a BLEU score of 22.04, and with two, 25.25.
As another example, the NIST OpenMT Arabic--English and Chinese--English test sets\footnote{\url{https://catalog.ldc.upenn.edu/LDC2010T21}} provided four references and consequently yielded BLEU scores in the high 40s (and now, low 50s).
Since these numbers are all gathered together under the label ``BLEU'', over time, they leave an impression in people's minds of very high BLEU scores for some language pairs or test sets relative to others, but without this critical distinguishing detail.

\subsection{Problem: Different reference preprocessings cannot be compared}
\label{section:processing}

The first problem dealt with parameters used in BLEU scores, and was more theoretical.
A second problem, that of preprocessing, exists in practice.

Preprocessing includes input text modifications such as normalization (e.g., collapsing punctuation, removing special characters), tokenization (e.g., splitting off punctuation), compound-splitting, the removal of case, and so on.
Its general goal is to deliver meaningful white-space delimited tokens to the MT system.
Of these, tokenization is one of the most important and central.
This is because BLEU is a precision metric, and changing the reference processing changes the set of n-grams against which system n-gram precision is computed.
\citet{rehbein2007:treebank} showed that the analogous use in the parsing community of F$_1$ scores as rough estimates of cross-lingual parsing difficulty were unreliable, for this exact reason.
BLEU scores are often reported as being \emph{tokenized} or \emph{detokenized}.
But for computing BLEU, both the system output and reference are always tokenized; what this distinction refers to is whether the reference preprocessing is \emph{user-supplied} or \emph{metric-internal} (i.e., handled by the code implementing the metric), respectively.
And since BLEU scores can only be compared when the reference processing is the same, user-supplied preprocessing is error-prone and inadequate for comparing across papers.

Table~\ref{table:scores} demonstrates the effect of computing BLEU scores with different reference tokenizations.
This table presents BLEU scores where a single WMT 2017 system (online-B) and the reference translation were both processed in the following ways:
\begin{itemize}
\item \emph{basic}.
  User-supplied preprocessing with the \moses tokenizer \citep{koehn2007:moses}.\footnote{Arguments \texttt{-q -no-escape -protected basic-protected-patterns -l LANG}.}
\item \emph{split}.
  Splitting compounds, as in \citet{luong2015:effective}:\footnote{Their use of compound splitting is not mentioned in the paper, but only here: \url{http://nlp.stanford.edu/projects/nmt}.} e.g., \emph{rich-text} $\rightarrow$ \emph{rich - text}.
\item \emph{unk}.
  All word types not appearing at least twice in the target side of the WMT training data (with ``basic'' tokenization) are mapped to UNK.
  This hypothetical scenario could easily happen if this common user-supplied preprocessing were inadvertently applied to the reference.
\item \emph{metric}.
  Only the metric-internal tokenization of the official WMT scoring script, \verb|mteval-v13a.pl|, is applied.\footnote{\url{https://github.com/moses-smt/mosesdecoder/blob/master/scripts/generic/mteval-v13a.pl}}
\end{itemize}

The changes in each column show the effect these different schemes have, as high as 1.8 for one arc, and averaging around 1.0.
The biggest is the treatment of case, which is well known, yet many papers are not clear about whether they report cased or case-insensitive BLEU.

Allowing the user to handle pre-processing of the reference has other traps.
For example, many systems (particularly before sub-word splitting \citep{sennrich2016:neural} was proposed) limited the vocabulary in their attempt to deal with unknown words.
It's possible that these papers applied this same unknown-word masking to the references, too, which would artificially inflate BLEU scores.
Such mistakes are easy to introduce in researcher pipelines.\footnote{This paper's observations stem in part from an early version of a research workflow I was using, which applied preprocessing to the reference, affecting scores by half a point.}

\subsection{Problem: Details are hard to come by}
\label{section:details}

User-supplied reference processing precludes direct comparison of published numbers, but if enough detail is specified in the paper, it is at least possible to reconstruct comparable numbers.
Unfortunately, this is not the trend, and even for meticulous researchers, it is often unwieldy to include this level of technical detail.
In any case, it creates uncertainty and work for the reader.
One has to read the experiments section, scour the footnotes, and look for other clues which are sometimes scattered throughout the paper.
Figuring out what another team did is not easy.

\begin{table}[t]
  \centering
  \begin{tabular}{l|l}
    paper & configuration \\
    \hline\hline
    \citet{chiang2005:hierarchical} & metric\lc
    \\
    \citet{bahdanau2014:neural} & \unclear
    \\
    \citet{luong2015:addressing} & user or metric \unclear
    \\
    \citet{jean2015:using} & user
    \\
    \citet{wu2016:googles} & user or user\lc \unclear
    \\
    \citet{vaswani2017:attention} & user or user\lc \unclear
    \\
    \citet{gehring:2017:convolutional} & user, metric
    \\
  \end{tabular}
  \caption{Benchmarks set by well-cited papers use different BLEU configurations (Table~\ref{table:scores}).
    Which one was used is often difficult to determine.}
  \label{table:survey}
\end{table}

The variations in Table~\ref{table:scores} are only some of the possible configurations, since there is no limit to the preprocessing that a group could apply.
But assuming these represent common, concrete configurations, one might wonder how easy it is to determine which of them was used by a particular paper.
Table~\ref{table:survey} presents an attempt to recover this information from a handful of influential papers in the literature.
Not only are systems not comparable due to different schemes, in many cases, no easy determination can be made.

\subsection{Problem: Dataset specification}
\label{section:wmt14}

Other tricky details exist in the management of datasets.
It has been common over the past few years to report results on the English$\rightarrow$German arc of the WMT'14 dataset.
It is unfortunate, therefore, that for this track (and this track alone), there are actually \emph{two} such datasets.
One of them, released for the evaluation, has only 2,737 sentences, having removed about 10\% of the original data after problems were discovered during the evaluation.
The second, released after the evaluation, restores this missing data (after correcting the problem) and has 3,004 sentences.
Many researchers are unaware of this fact, and do not specify which version they use when reporting, which itself contributes to variance.

\subsection{Summary}

\begin{figure}[t]
  \begin{center}
    \includegraphics[width=50mm]{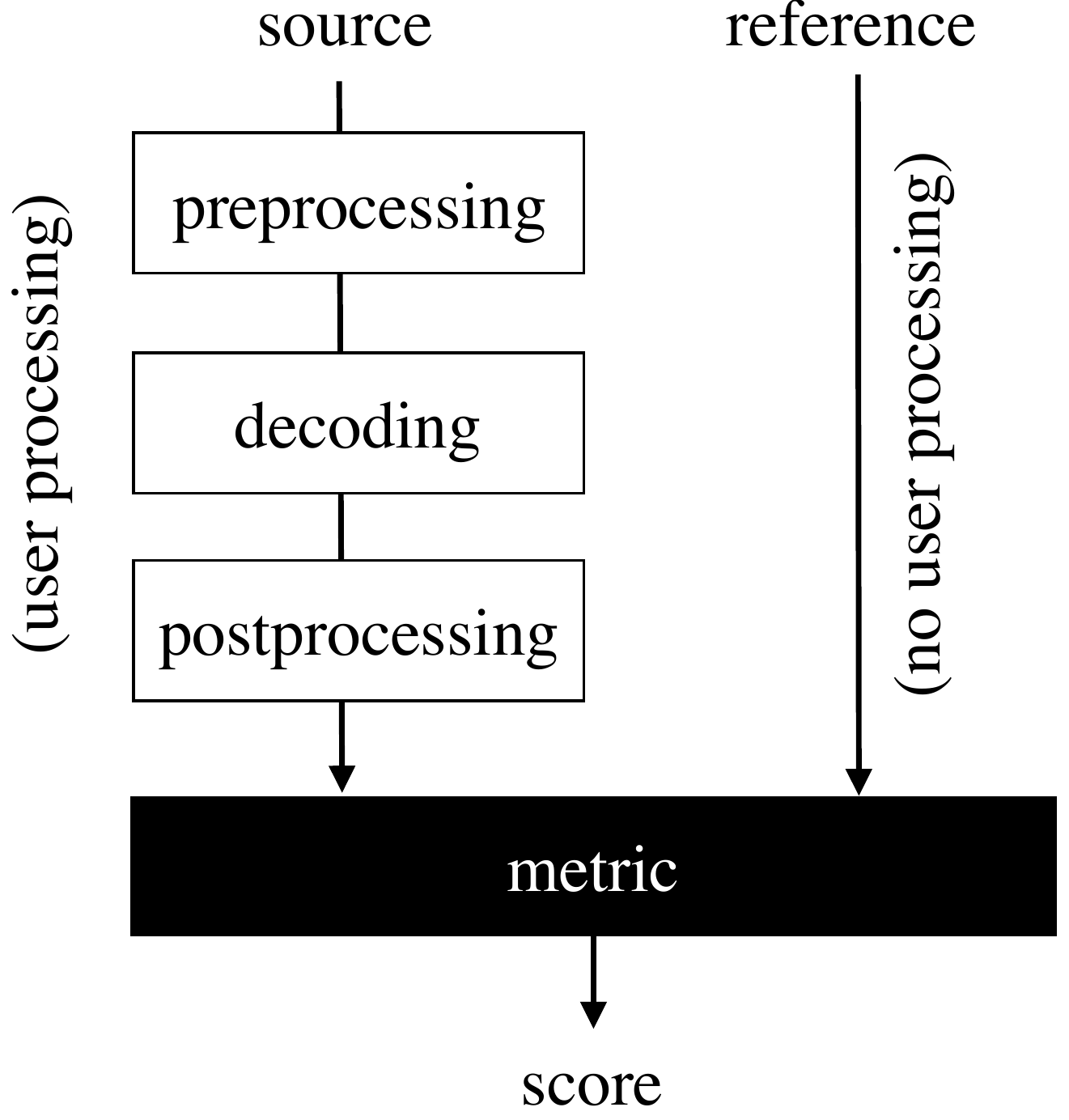}
  \end{center}
  \caption{The proper pipeline for computing reported BLEU scores.
    White boxes denote user-supplied processing, and the black box, metric-supplied.
    The user should not touch the reference, while the metric applies its own processing to the system output and reference.}
  \label{figure:pipeline}
\end{figure}

Figure~\ref{figure:pipeline} depicts the ideal process for computing sharable scores.
Reference tokenization must identical in order for scores to be comparable.
The widespread use of user-supplied reference preprocessing prevents this, needlessly complicating comparisons.
The lack of details about preprocessing pipelines exacerbates this problem.
This situation should be fixed.

\section{A way forward}

\subsection{The example of PARSEVAL}

An instructive comparison comes from the evaluation of English parsing scores, where numbers have been safely compared across papers for decades using the PARSEVAL metric \citep{black1991:procedure}.
PARSEVAL works by taking labeled spans of the form $(N,i,j)$ representing a nonterminal $N$ spanning a constituent from word $i$ to word $j$.
These are extracted from the parser output and used to compute precision and recall against the gold-standard set taken from the correct parse tree.
Precision and recall are then combined to compute the F$_1$ metric that is commonly reported and compared across parsing papers.

Computing parser F$_1$ comes with its own set of hidden parameters and edge cases.
Should one count the \verb|TOP| (\verb|ROOT|) node?
What about \verb|-NONE-| nodes?
Punctuation?
Should any labels be considered equivalent?
These boundary cases are resolved by that community's adoption of a standard codebase, \verb|evalb|,\footnote{\url{http://nlp.cs.nyu.edu/evalb/}} which included a parameters file that answers each of these questions.\footnote{The configuration file, \texttt{COLLINS.PRM}, answers these questions as no, no, no, and ADVP=PRT.}
This has facilitated almost thirty years of comparisons on treebanks such as the Wall Street Journal portion of the Penn Treebank \citep{marcus1993:building}.

\subsection{Existing scripts}

\moses\footnote{\url{http://statmt.org/moses}} has a number of scoring scripts.
Unfortunately, each of them has problems.
Moses' \verb|multi-bleu.perl| cannot be used because it requires user-supplied preprocessing.
The same is true of another evaluation framework, MultEval \citep{clark2011:better}, which explicitly advocates for user-supplied tokenization.\footnote{\url{https://github.com/jhclark/multeval}}
A good candidate is Moses' \verb|mteval-v13a.pl|, which makes use of metric-internal preprocessing and is used in the annual WMT evaluations.
However, this script inconveniently requires the data to be wrapped into XML.
Nematus \cite{sennrich2017:nematus} contains a version (\verb|multi-bleu-detok.perl|) that removes the XML requirement.
This is a good idea, but it still requires the user to manually handle the reference translations.
A better approach is to keep the reference away from the user entirely.

\subsection{\sacrebleu}

\sacrebleu is a Python script that aims to treat BLEU with a bit more reverence:

\begin{itemize}
\item It expects detokenized outputs, applying its own metric-internal preprocessing, and produces the same values as WMT;
\item it automatically downloads and stores WMT (2008--2018) and IWSLT 2017 \citep{cettolo2017:overview} test sets, obviating the need for the user to handle the references at all; and
\item it produces a short version string that documents the settings used.
\end{itemize}

\sacrebleu can be installed via the Python package management system:

\begin{verbatim}
  pip3 install sacrebleu
\end{verbatim}

It can then be used to download the source side of test sets as decoder input---all WMT test sets are available, as well as recent IWSLT test sets, and others are being added.
After decoding and detokenization, it can then used to produce BLEU scores.\footnote{The CHRF metric is also available via the \texttt{-m} flag.}
The following command selects the WMT'14 EN-DE dataset used in the official evaluation:
\begin{verbatim}
  cat output.detok \
    | sacrebleu -t wmt14 -l en-de
\end{verbatim}
(The restored version that was released after the evaluation (\S\ref{section:wmt14}) can be selected by using \verb|-t wmt14/full|.)
It prints out a version string recording all the parameters as '+' delimited KEY.VALUE pairs (here shortened with \verb|--short|):
\begin{verbatim}
  BLEU+c.mixed+l.en-de+#.1+s.exp
    +t.wmt14+tok.13a+v.1.2.10
\end{verbatim}
recording:
\begin{itemize}
\item mixed case evaluation
\item on EN-DE
\item with one reference
\item and exponential smoothing
\item on the WMT14 dataset
\item using the WMT standard '13a' tokenization
\item with \sacrebleu \sacrebleuversion.
\end{itemize}

\sacrebleu is open source software released under the Apache 2.0 license.

\section{Summary}

Research in machine translation benefits from the regular introduction of test sets for many different language arcs, from academic, government, and industry sources.
It is a shame, therefore, that we are in a situation where it is difficult to directly compare scores across these test sets.
One might be tempted to shrug this off as an unimportant detail, but as was shown here, these differences are in fact quite important, resulting in large variances in the score that are often much higher than the gains reported by a new method.

Fixing the problem is relatively simple.
Research groups should only report BLEU computed using a metric-internal tokenization and preprocessing scheme for the reference, and they should be explicit about the BLEU parameterization they use.
With this, scores can be directly compared.
For backwards compatibility with WMT results, I recommend the processing scheme used by WMT, and provide a new tool that makes it easy to do so.

\bibliography{main}
\bibliographystyle{latex/acl_natbib_nourl}

\end{document}